\documentclass[manuscript, screen, nonacm]{acmart}

\usepackage{rotating}

\usepackage{booktabs}
\usepackage{siunitx}
\usepackage{makecell}
\usepackage[table]{xcolor}

\setcopyright{acmlicensed}
\copyrightyear{2018}
\acmYear{2018}
\acmDOI{XXXXXXX.XXXXXXX}
\acmConference[Conference acronym 'XX]{Make sure to enter the correct
	conference title from your rights confirmation email}{June 03--05,
	2018}{Woodstock, NY}
\acmISBN{978-1-4503-XXXX-X/2018/06}

\begin{document}

\title{Eye image segmentation using visual and concept prompts with Segment Anything Model 3 (SAM3)}

\author{Diederick C. Niehorster}
\email{diederick_c.niehorster@humlab.lu.se}
\affiliation{%
	\institution{Lund University Humanities Lab \& Dept. of Psychology, Lund University}
	\city{Lund}
	\country{Sweden}}
\orcid{0000-0002-4672-8756}

\author{Marcus Nyström}
\email{marcus.nystrom@humlab.lu.se}
\affiliation{%
	\institution{Lund University Humanities Lab}
	\city{Lund}
	\country{Sweden}
}

\renewcommand{\shortauthors}{Niehorster et al.}

\begin{abstract}
	Previous work has reported that vision foundation models show promising zero-shot performance in eye image segmentation. Here we examine whether the latest iteration of the Segment Anything Model, SAM3, offers better eye image segmentation performance than SAM2, and explore the performance of its new concept (text) prompting mode. Eye image segmentation performance was evaluated using diverse datasets encompassing both high-resolution high-quality videos from a lab environment and the TEyeD dataset consisting of challenging eye videos acquired in the wild. Results show that in most cases SAM3 with either visual or concept prompts did not perform better than SAM2, for both lab and in-the-wild datasets. Since SAM2 not only performed better but was also faster, we conclude that SAM2 remains the best option for eye image segmentation. We provide our adaptation of SAM3's codebase that allows processing videos of arbitrary duration.
\end{abstract}

\ccsdesc[500]{Computing methodologies~Video segmentation}
\ccsdesc[300]{Human-centered computing}
\ccsdesc[300]{Computing methodologies~Neural networks}

\keywords{Eye tracking, Feature localization, Gaze estimation, Foundation models, Prompting, Methods, Pupil, Corneal reflection, Iris, Sclera}
%

\maketitle

\section{Introduction}

There are several approaches to performing gaze estimation, that is, determining where a person looks based on images from their eyes \citep{liu2022survey}. Besides appearance-based methods that often rely on end-to-end neural networks \citep{cheng2024appearancebased}, most other approaches require features, such as the center or edges of the pupil, the iris and the corneal reflection (CR) of the eye tracker's illuminator, to be detected and localized in the eye images. Such features are then used for regression-based gaze estimation techniques such as P-CR \citep[e.g.,][]{merchant1974remote,stampe1993heuristic,kliegl1981calib,cerrolaza2012study,blignaut2013effect} or techniques involving geometric eye models [\citealp{guestrinEizenman2006,coutinho2006headfree,barsingerhorn2017model,swirski2013fully,dierkes2018novel,santini2019grip}; see \citealp{hansen2010eye} for an overview]. How are these features detected in eye images and their centers localized? While traditional image processing approaches to detecting these features \citep[e.g.,][]{nystrom2023amplitude,swirski2012robust,santini2018purest,fuhl2015excuse,fuhl2016else} remain popular, deep learning approaches are also being developed [\citealp{fuhl2017pupilnet,fuhl2020tiny,fuhl2023pistol,kothari2021ellseg,deng2024sam,cheng2024appearancebased,kim2019nvgaze}; see \citealp{akinyelu2020convolutional} for an overview]. One recurring problem with deep learning approaches is that if they do not work out of the box, retraining requires large datasets that are expensive to acquire and manually annotate \citep[e.g.,][]{byrne2023CRCNN,byrne2025leyes}. Vision foundation models, such as the Segment Anything Model family of models \citep{kirillov2023sam,ravi2024sam2,carion2025sam3}, offer a potential solution to eye image segmentation that may drastically reduce the amount of manual work required. Specifically, both SAM \citep{maquiling2024sam,deng2024sam} and SAM 2 \citep{maquiling2025sam2,niehorster2025sam2} have been shown to offer segmentation performance that is competitive with the state-of-the-art for eye images in a range of different settings. The eye image segmentation performance of the latest model, SAM 3 \citep{carion2025sam3} has however not yet been examined.

In this work, we examine how the recently released SAM 3 performs compared to SAM 2 for the task of eye image segmentation. Specifically, first we ask whether the visual prompting (manually indicating an object in the image to be segmented) performance of SAM 3 is superior to that of SAM 2. While SAM 3 has the same model architecture as SAM 2 for visual prompting tasks, the HieraDet image encoder \citep{ryali2023hiera,bolya2023windowattention} of SAM 2 was replaced with the Perception Encoder \citep{bolya2025perceptionencoder}, which may lead to altered performance. Second, we ask how SAM 3's new concept prompting abilities (i.e., prompting the model using a short string such as ``pupil'') compares to visual prompting in SAM 2 and 3. Concept prompting would remove all manual work from using SAM (no prompts have to be placed manually on the various eye features), but it also offers potential benefits for, for instance, blink recovery.

To provide insight into how SAM 2 and SAM 3 perform in eye image segmentation across a wide range of eye tracking domains, we perform our evaluation using both datasets containing high-resolution and high-quality eye images obtained in controlled lab settings, and datasets of eye images obtained from head-worn (glasses, VR and AR) eye trackers obtained in unconstrained settings. Across datasets, we prompt the models to segment the pupil, iris and sclera, and for the high-resolution lab datasets we additionally segment the CR. For the lab datasets, since they lack a ground truth, we examine the RMS-S2S \citep{niehorster2020characterizing,niehorster2020impact} precision of the resulting feature signals, thereby probing which method yields signals with the lowest noise levels, and data loss. For the in-the-wild dataset, TEyeD \citep{fuhl2021teyed} is used, which consists of various datasets \citep[][]{tonsen2016labelled,kothari2020gaze,kim2019nvgaze,fuhl2015excuse,fuhl2016else,fuhl2016pupilnet,kasneci2014driving} and ground truth labels for the pupil, iris and eyelid aperture. As such, for these datasets, model performance is assessed as the overlap of model output with the ground truth segmentation, using the intersection-over-union (IoU) metric. Finally, as part of this work, we have adapted the SAM 3 code to be able to run on videos of arbitrary length. This code is available here: \url{https://github.com/dcnieho/sam3}.


\section{Methods}
\subsection{Datasets}
\subsubsection{High-resolution lab datasets}
Two different datasets were used, both recorded using the FLEX setup \citep{hooge2021,nystrom2023amplitude,byrne2023CRCNN,byrne2025leyes,hooge2024eyehead,valtakari2024fieldtest}. The first dataset was recorded from four experienced participants (all male) at 1000 Hz and consisted of 106 trials during which participants performed various short and long fixation and saccade sequences \citep[this dataset has been used previously, see][for further description]{byrne2025leyes,niehorster2025sam2}. The second dataset consisted of recordings taken from 17 participants (five females, 11 males, one non-binary). Each dataset contained forty trials consisting of fixation and saccade sequences recorded at 1000 Hz \citep[the dataset was previously used, see][for further details]{byrne2023CRCNN}. The datasets contain 2.87 million images.

\subsubsection{Unconstrained datasets}
TEyeD \citep{fuhl2021teyed} consists of NVGaze \citep{kim2019nvgaze}, two sets of eye images collected with a VR (14 participants) and an AR (42 participants) device; Gaze-in-wild \citep{kothari2020gaze}, 19 participants performing everyday tasks with a wearable eye tracker; Labelled pupils in the wild \citep{tonsen2016labelled}, 22 participants performing everyday tasks with a wearable eye tracker; and a series of Dikablis datasets, which are a combination of the datasets from \citet{fuhl2015excuse}, \citet{fuhl2016else}, \citet{fuhl2016pupilnet} and \citet{kasneci2014driving}. TEyeD provides ground truth pupil and iris ellipses as well as eyelid polygons for all datasets. The datasets contain 14.44 million images.

\subsection{Models and prompting}
\begin{figure*}[h]
	\centering
	\includegraphics[width=\linewidth]{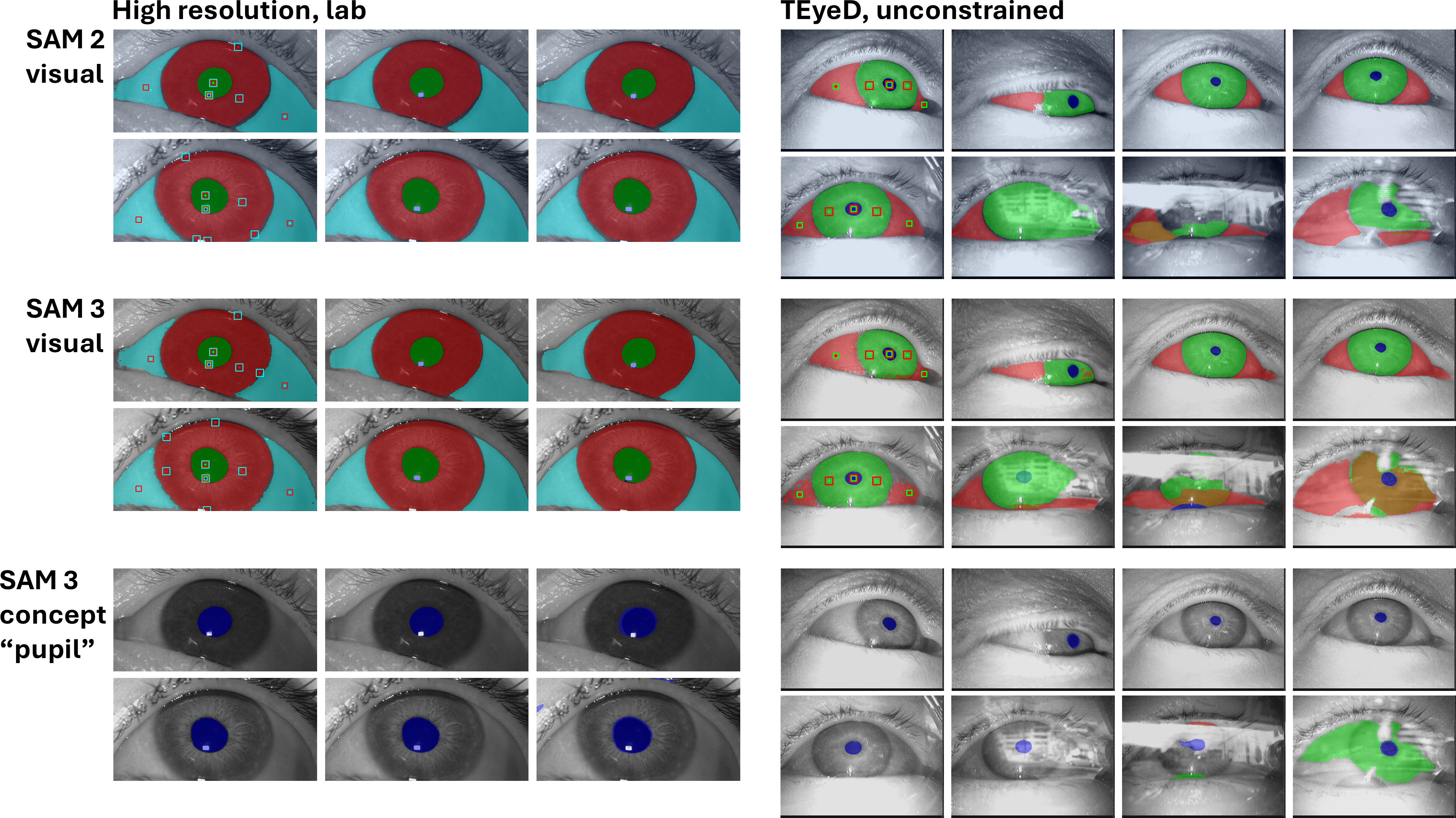}
	\caption{Example prompts and resulting segmentations for the high resolution lab datasets (left) and the TEyeD datasets (right) for SAM 2 visual prompting (top), SAM 3 visual prompting (middle) and SAM 3 concept prompting (bottom). For the lab images, the prompted frame, the 100th frame and the 1000th frame in the eye video are shown. For the TEyeD images, the prompted frame, the 1000th, the 10000th and the 20000th frame are shown. For the lab datasets (left): the dark blue mask indicates the CR, green the pupil, red the iris and cyan the sclera. For the TEyeD datasets (right): dark blue is the pupil, green the iris, red the sclera. Positive (+) and negative (square) prompts use the same color codes. For the bottom row, different colors instead indicate different ``pupil'' objects returned by the model. The brown in some segmentations results from the iris and sclera masks overlapping.}
	\Description{Close-up eye images with different segmentations of parts of the eye.}
	\label{fig:examples}
\end{figure*}

For SAM 2 \citep{ravi2024sam2}, the largest model (\texttt{sam2.1\_hiera\_large}) was used since it performs best for eye image segmentation \citep{niehorster2025sam2}. For SAM 3 \citep{carion2025sam3}, only one model size is available (\texttt{sam3.pt} from 19 Nov 2025). Below, we describe how prompting was performed, Figure \ref{fig:examples} shows example prompts and segmentation masks. Inference was run on three computer systems, (A) a 64-core AMD Threadripper Pro 9985WX system with 1TB of memory and an nVidia RTX Pro 6000 Blackwell GPU (96GB VRAM), (B) a 32-core AMD Threadripper 3970X system with 128GB of memory and an nVidia RTX 4090 GPU (24GB VRAM, Ada Lovelace architecture) and (C) a 12-core Intel i9-9920X system with 64GB of memory and two nVidia Titan RTX GPUs (24GB VRAM, Turing architecture), all running on Windows 10 or 11. SAM 2 with visual prompts consumed about 10GB of VRAM and ran at 16 fps on system (A), 13 fps on (B) and 1.6 fps on (C). SAM 3 with visual prompts ran at 13 fps on system (A), 9 fps on (B) and 0.6 fps on (C). It consumed 7.5GB of VRAM on (A) and (B), but 20GB on (C). SAM 3 with concept prompts was only run on (A), consumed 8GB of VRAM, and ran at 11 fps initially, but slowed down to 0.5 fps for longer (e.g. 70000 frames) videos.

\subsubsection{Visual prompts}
For the high-resolution datasets, the refined prompting strategy of \citet{niehorster2025sam2} was used. For each participant, a single image taken when the participant was looking at the middle of the screen was used to prompt all videos recorded for that participant. On this image, one prompt was manually placed by the first author on the CR, one on the pupil, one on the iris and two on the sclera (one on each side of the iris). Additionally, the CR prompt also served as a negative prompt for the pupil, iris and sclera (indicating that the CR should not be part of the segmentation for these features); the pupil as a negative prompt for the iris and sclera, the iris as negative prompt for the sclera, and the two sclera prompts as negative prompts for the iris. The segmentation output of both SAM 2 and SAM 3 for these initial prompt sets was then examined and, per model and prompt image, additional positive and negative prompts were added manually until segmentation was judged satisfactory by the annotator. For all prompt images, this only involved additional iris and sclera prompts. Any new positive iris or sclera prompts were also added as negative prompts for the sclera or iris, respectively.

\begin{figure*}[h]
	\centering
	\includegraphics[width=.65\linewidth]{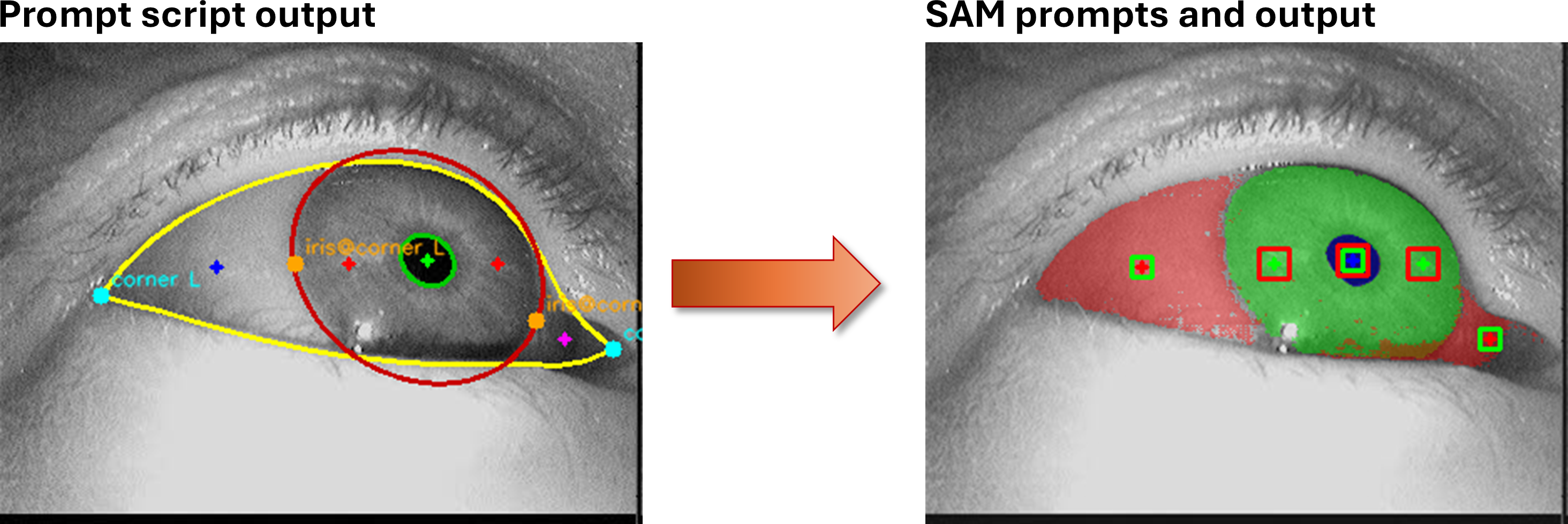}
	\caption{Prompting TEyeD. The left panel shows the ground truth annotations provided by TEyeD for the pupil (green ellipse), iris (red ellipse) and palpebral fissure (yellow polygon). Also indicated are the determined eye corners (cyan dots), closest points on the iris ellipse (orange points) and the derived prompt coordinates for the pupil (green +), the iris (red +) and the sclera (blue and magenta +). The right panel shows the corresponding positive (+) and negative (square) visual prompts provided to both SAM models (blue: pupil, green: iris, red: sclera). Also shown are the output masks from SAM 3, using the same color codes as the prompts.}
	\Description{Close-up eye images with ground truth annotation and output segmentation of parts of the eye.}
	\label{fig:teyed_prompting}
\end{figure*}

For the TEyeD datasets, the minimal prompting strategy of \citet{niehorster2025sam2} was implemented using a script (available from \textbf{<link blinded for review>}), as it was not feasible to manually create a refined prompt for each eye video due to the large number of videos. First, a frame suitable for prompting, i.e. where the eye is sufficiently open, was manually selected for each eye video. Then, using the ground truth annotations for the pupil, the iris and the eyelids, a prompt set was created as follows (see Figure \ref{fig:teyed_prompting}). First, one positive prompt was placed on the pupil and two on the iris (one on each side of the pupil) such that the prompt was inside the eyelid polygon with a margin of at least 10 pixels. Then, two positive prompts were placed on the sclera (one on each side of the iris) using the following logic. First, the eye corners and the closest point on the iris ellipse that is inside the eyelid polygon are found. Then a point is placed at 40\% along the line from the point on the iris border to the eye corner. A perpendicular line is drawn through this point and intersected with the eyelid polygon. The final prompt is placed on this perpendicular line, in the middle between the two points where it intersects the eyelid polygon.

\subsubsection{Concept prompts} In an initial run using several video files, the strings ``pupil'', ``iris'' or ``sclera'' were used as concept prompts. Since no segmentation was returned for the iris and sclera, these prompts were not used for the other video files. As such, only results for ``pupil'' are reported in this paper. We will return to this issue in the discussion.

\subsection{Data analysis}
For the high-resolution datasets, first, feature signals were constructed from the output masks of the models. For the CR and pupil, feature centers were determined as the center of mass of the contour with the largest area in the output masks. For the output of SAM 3 with concept prompts, since it may return multiple objects, after applying shape criteria, the blob closest to the center of the image is selected for the first frame. For subsequent frames a simple tracking algorithm is used that selects the blob closest to the blob from the previous frame. If track is lost, the blob closest to the center is again used to reinitialize tracking. For the iris, the method of \citet{niehorster2025sam2} was used, which employs the sclera output mask to determine which edges of the iris output mask are not adjacent to the eyelid and then fits an ellipse to this part of the iris borders. The iris signal then consists of the center of the fitted ellipse. We assess the quality of the CR, pupil and iris feature signals using data loss and, following previous work \citep{byrne2023CRCNN,byrne2025leyes,niehorster2025sam2}, RMS-S2S precision \citep{niehorster2020impact,niehorster2020characterizing,holmqvist2012edq}. RMS-S2S precision was computed using a moving 200-ms window, and the precision for a given video then determined as the median RMS value across these windows \citep{hooge2018goldenstandard,niehorster2020glassesviewer,hooge2022test}. Data loss was determined separately for the CR, pupil, iris and sclera from the output masks of the models. To determine data loss for each of these eye parts, due to blinks or otherwise, first for each video the distribution of areas of the masks for each eye part were determined. Data loss was then flagged for a given frame and object when the area was less than half of the 20th percentile value. Given how the output masks were processed for creating the feature signals (see above), for the CR and pupil, the area of the largest contour in the mask is used. For the iris and sclera instead the total number of pixels in the model's output masks were used. Differences in performance between the models are assessed using paired t-tests.

For the TEyeD datasets, for each frame in each video, the Intersection-over-Union (IoU, ranging from 0 [no overlap] to 1 [perfect overlap]) of the model and ground truth were determined. TEyeD provides ground truth pupil ellipses, iris ellipses and eyelid polygons, the latter annotating the palpebral fissure. Since the ground truth pupil and iris ellipses include parts of these features occluded by the eyelid, the visible part of the ground truth pupil or iris is first determined by intersecting them with the eyelid ground truth. For the iris mask, the pupil was also first removed as it should not be part of the iris output mask given the used prompts. To compute IoU for the sclera, the ground truth pupil and iris are removed from the eyelid ground truth. We also computed IoU for the entire eye opening, using the union of the pupil, iris and sclera output masks provided by the model. For SAM 2 and SAM 3 with visual prompts, the output masks were used directly to compute IoU. Since SAM 3 with concept prompts could return multiple ``pupil'' objects (see Figure \ref{fig:examples}), we performed the same selection and tracking logic as for the high-resolution lab datasets described above. Besides IoU, several other metrics were computed. First, since we noted that the sclera mask often includes the iris despite the negative prompts placed on the iris, to further assess the quality of the output sclera masks we computed how much of the iris mask was also included in the sclera mask. Furthermore, for the pupil, iris and sclera we determined the false alarm rate (when the feature is not present in the ground truth but reported by the model) and the miss rate (feature present in ground truth but not in the model output). Finally, to assess the overall ability of the models to correctly classify feature presence, we use Youden's J \citeyearpar{youden1950j}. Youden's J incorporates both the hit rate (1--miss rate) and false alarm rate to assess the trade-off between a model's sensitivity and specificity. Specifically, if the model has a very low miss rate but also a very high false alarm rate (i.e., it indiscriminately reports that a feature is present), Youden's J is zero, while perfect performance (both miss rate and false alarm rate are 0) is indicated by a Youden's J of 1.

\section{Results}

\begin{figure*}[h]
	\centering
	\includegraphics[width=\linewidth]{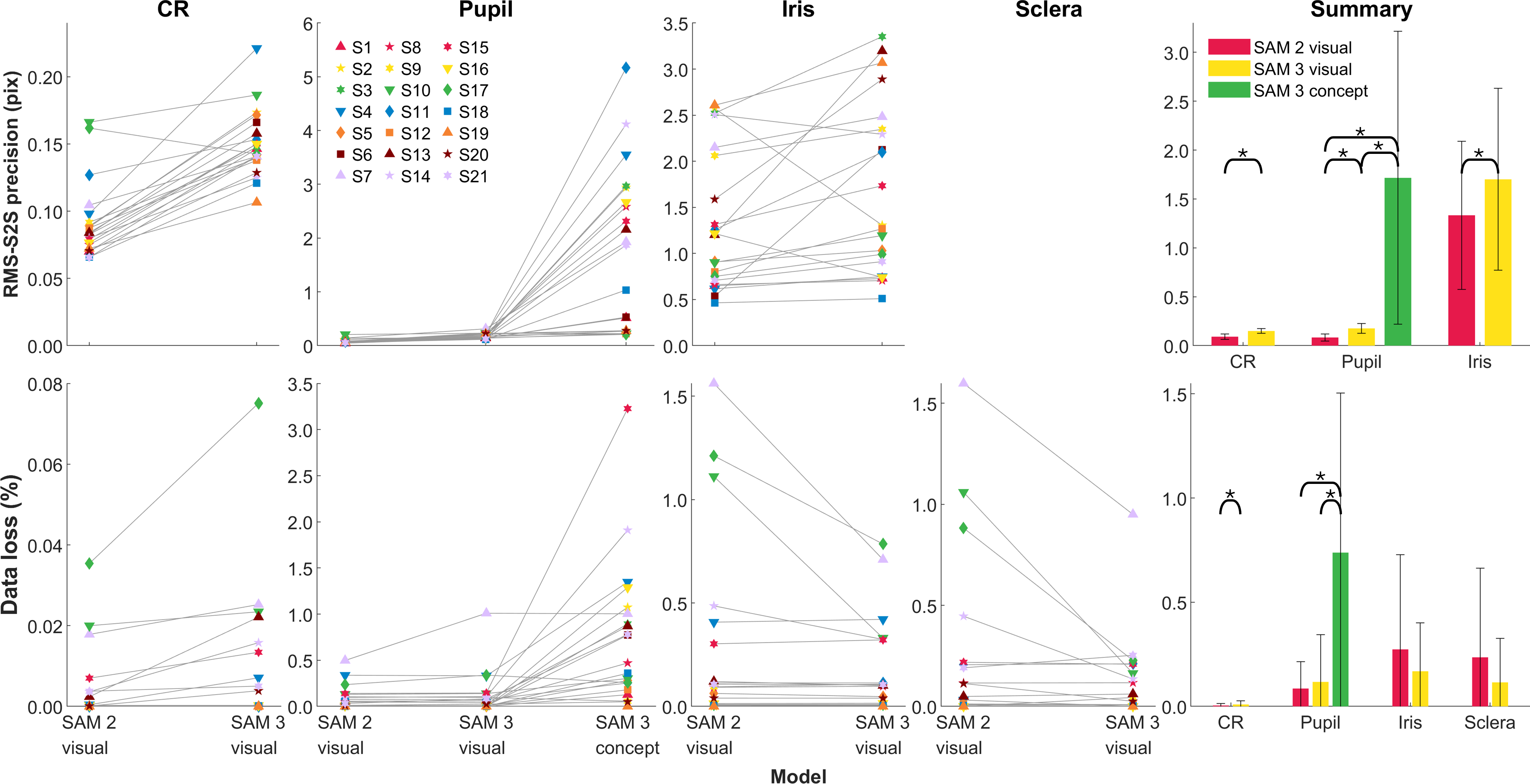}
	\caption{RMS-S2S precision and data loss for SAM 2 and SAM 3 on the high-resolution lab datasets. Shown are the RMS-S2S precision (lower values is better) and data loss (lower is better) per participant. Note that the range of the y-axis is different for each of the panels. A summary showing the precision or data loss on the same scale is shown in the right most bar graphs (error bars indicate SEM across participants). Stars indicate significant differences according to paired t-tests.}
	\Description{A collection of line and bar graphs.}
	\label{fig:result_lab}
\end{figure*}

Figure \ref{fig:result_lab} (top row) shows the RMS-S2S precision achieved by SAM 2 with visual prompts and SAM 3 with visual or concept prompts. As can be seen, while there was significant between-participant variability, within participants, RMS-S2S precision was systematically worse for SAM 3 with visual prompts than for SAM 2 with visual prompts. In fact, the average RMS-S2S precision was 64\% worse for the CR, 112\% worse for the pupil and 28\% for the iris. The RMS-S2S precision of the pupil center signal derived from the segmentation masks produced by SAM 3 with concept prompting was much worse than for either SAM 3 with visual prompts (875 \%) or for SAM 2 (1966\%). Data loss (Figure \ref{fig:result_lab}, bottom row) was low overall, given that there were very few blinks in the high-quality datasets and that segmentation was successful. For the most part, differences between the three models were also small. Nonetheless, for the CR and pupil, data loss was lower for SAM 2 with visual prompts than for SAM 3 with visual prompts (111\% and 37\%, respectively). Data loss for the iris and sclera was not singificantly different between SAM 2 and 3. Data loss for the pupil for SAM 3 with concept prompts was 528\% higher than SAM 3 with visual prompts and 762 \% higher than for SAM 2.

\begin{table}[ht]
	\centering
	\setlength{\tabcolsep}{3pt}
	\begin{tabular}{l S[table-format = 1.3] S[table-format = 1.3] S[table-format = 1.3] S[table-format = 1.3] !{\color{black!20}\vrule width 0.3pt} S[table-format = 1.3] !{\color{black!20}\vrule width 0.3pt} S[table-format = 1.3] S[table-format = 1.3] S[table-format = 1.3] !{\color{black!20}\vrule width 0.3pt} S[table-format = 1.3] S[table-format = 1.3] S[table-format = 1.3] !{\color{black!20}\vrule width 0.3pt} S[table-format = 1.3] S[table-format = 1.3] S[table-format = 1.3]}
		\toprule
		Dataset & \multicolumn{4}{c}{\makecell{mIoU}} & \multicolumn{1}{c}{\makecell{Overlap}} & \multicolumn{3}{c}{\makecell{FA rate}} & \multicolumn{3}{c}{\makecell{Miss rate}} & \multicolumn{3}{c}{\makecell{Youden's J}} \\
		\cmidrule(lr){2-5} \cmidrule(lr){6-6} \cmidrule(lr){7-9} \cmidrule(lr){10-12} \cmidrule(lr){13-15}
		& \multicolumn{1}{c}{\makecell{pupil}} & \multicolumn{1}{c}{\makecell{iris}} & \multicolumn{1}{c}{\makecell{sclera}} & \multicolumn{1}{c}{\makecell{eye\\opening}} & \multicolumn{1}{c}{\makecell{iris--sclera}} & \multicolumn{1}{c}{\makecell{Pupil}} & \multicolumn{1}{c}{\makecell{Iris}} & \multicolumn{1}{c}{\makecell{Sclera}} & \multicolumn{1}{c}{\makecell{Pupil}} & \multicolumn{1}{c}{\makecell{Iris}} & \multicolumn{1}{c}{\makecell{Sclera}} & \multicolumn{1}{c}{\makecell{Pupil}} & \multicolumn{1}{c}{\makecell{Iris}} & \multicolumn{1}{c}{\makecell{Sclera}} \\
		\midrule
		\multicolumn{15}{c}{\textbf{SAM 2 visual}} \\
		Dikablis & \bfseries 0.880 & \bfseries 0.814 & 0.495 & \bfseries 0.773 & 0.270 & \bfseries 0.065 & \bfseries 0.384 & \bfseries 0.377 & 0.037 & 0.003 & 0.011 & \bfseries 0.898 & \bfseries 0.613 & \bfseries 0.612 \\
		GiW & \bfseries 0.911 & \bfseries 0.768 & 0.464 & \bfseries 0.767 & 0.483 & \bfseries 0.193 & \bfseries 0.340 & \bfseries 0.373 & \underline{0.010} & 0.032 & 0.050 & \bfseries 0.798 & \bfseries 0.628 & \bfseries 0.578 \\
		LPW & \bfseries 0.853 & 0.692 & 0.333 & 0.639 & 0.587 & \underline{0.506} & \bfseries 0.668 & \bfseries 0.851 & \underline{0.023} & 0.055 & 0.073 & \underline{0.470} & \bfseries 0.278 & 0.076 \\
		NVGaze & \bfseries 0.895 & \bfseries 0.854 & 0.593 & \bfseries 0.816 & 0.227 & \bfseries 0.195 & \bfseries 0.447 & \bfseries 0.548 & \underline{0.011} & 0.008 & 0.009 & \bfseries 0.794 & \bfseries 0.545 & \bfseries 0.443 \\
		\midrule\addlinespace[0.6em]
		\multicolumn{15}{c}{\textbf{SAM 3 visual}} \\
		Dikablis & 0.782 & 0.745 & \bfseries 0.507 & 0.766 & \bfseries 0.202 & 0.394 & 0.793 & 0.892 & \underline{0.014} & \bfseries 0.002 & \bfseries 0.001 & 0.592 & 0.205 & 0.107 \\
		GiW & \underline{0.855} & 0.728 & \bfseries 0.506 & 0.752 & \bfseries 0.246 & 0.442 & 0.720 & 0.806 & \bfseries 0.006 & \bfseries 0.004 & \bfseries 0.003 & 0.552 & 0.276 & 0.190 \\
		LPW & \underline{0.800} & \bfseries 0.743 & \bfseries 0.425 & \bfseries 0.680 & \bfseries 0.307 & 0.676 & 0.893 & 0.914 & \bfseries 0.005 & \bfseries 0.002 & \bfseries 0.004 & 0.319 & 0.105 & \bfseries 0.082 \\
		NVGaze & \underline{0.884} & 0.843 & \bfseries 0.637 & 0.812 & \bfseries 0.093 & \underline{0.288} & 0.554 & 0.600 & \bfseries 0.006 & \bfseries 0.000 & \bfseries 0.000 & \underline{0.706} & 0.446 & 0.400 \\
		\midrule\addlinespace[0.6em]
		\multicolumn{15}{c}{\textbf{SAM 3 concept}} \\
		Dikablis & \underline{0.794} & \textemdash & \textemdash & \textemdash & \textemdash & \underline{0.301} & \textemdash & \textemdash & \bfseries 0.010 & \textemdash & \textemdash & \underline{0.689} & \textemdash & \textemdash \\
		GiW & 0.788 & \textemdash & \textemdash & \textemdash & \textemdash & \underline{0.244} & \textemdash & \textemdash & 0.025 & \textemdash & \textemdash & \underline{0.731} & \textemdash & \textemdash \\
		LPW & 0.784 & \textemdash & \textemdash & \textemdash & \textemdash & \bfseries 0.298 & \textemdash & \textemdash & 0.086 & \textemdash & \textemdash & \bfseries 0.617 & \textemdash & \textemdash \\
		NVGaze & 0.763 & \textemdash & \textemdash & \textemdash & \textemdash & 0.295 & \textemdash & \textemdash & 0.011 & \textemdash & \textemdash & 0.694 & \textemdash & \textemdash \\
		\bottomrule
	\end{tabular}
	\caption{Results for SAM 2 and SAM 3 per dataset in the TEyeD set. Shown are mIoU for the pupil, iris, sclera and the whole eye opening. Furthermore shown is the overlap between the iris and the sclera, and the false alarm rate, miss rate and Youden's J for the pupil, iris and sclera. For each column and dataset, the best values are printed in bold (higher is better for mIoU and Youden's J, while lower is better for the other columns), and for pupil columns the second best value is underlined.}
	\label{tab:teyed}
\end{table}

Table \ref{tab:teyed} lists the performance for the three models on the TEyeD datasets. The mean IoU (mIoU) was highest for SAM 2 for all datasets for the pupil and three out of the four datasets for the iris and entire eye opening. In contrast, for the sclera, mIoU was higher for SAM 3 with visual prompting than for SAM 2 for all four datasets. This may reflect that the amount of the iris output mask that is included in the sclera output mask (see the overlap column in Table \ref{tab:teyed}) was consistently lower for SAM 3 than for SAM 2. While SAM 3 with concept prompts performed worst of the three models for three out of the four datasets, it is not far behind the mIoU achieved by SAM 3 for the pupil with visual prompts. Examining the false alarm rate, it is seen that SAM 2 consistently outperforms SAM 3 with visual prompts for all data sets for the pupil, iris and sclera, while SAM 3 with concept prompts performs better than SAM 2 only for the pupil in one dataset. For the miss rate, however, SAM 3 outperforms SAM 2, with either SAM 3 with visual prompts (three datasets for the pupil and all datasets for the iris and sclera) or SAM 3 with concept prompts (one dataset for the pupil) showing a lower miss rate than SAM 2. However, the very low miss rate of SAM 3 is coupled with a very high false alarm rate, especially for the iris and sclera, indicating that the model often reports these features are present, regardless of whether they actually are. Indeed, Youden's J reveals that SAM 3 is (much) less discriminate for whether a feature is present in an eye image than SAM 2, especially for the iris and sclera.

\section{Discussion}
We examined the eye image segmentation performance of SAM 2 and SAM 3 on a diverse range of eye images, representing both a high-quality lab-based setting and unconstrained head-worn settings recorded in the wild. Across settings we found that, by and large, SAM 2 performed better than both modes of SAM 3, i.e., using either visual or concept prompts. Specifically, in the lab-based setting, the feature signals computed from SAM 2's segmentation were consistently less noisy than those computed from the segmentation provided by SAM 3 with visual prompts. SAM 3 with concept prompts performed, on average, an order of magnitude worse than SAM 2 or SAM 3 with visual prompts. No large differences in data loss were observed between the models. Nonetheless, data loss was lower for SAM 2 than SAM 3 with visual prompts for the CR and pupil, but not the iris or sclera. In the unconstrained setting evaluated using the TEyeD \citep{fuhl2021teyed} datasets, SAM 2 outperformed both SAM 3 models across most datasets in terms of mIoU for the pupil, iris and entire eye opening and also in terms of false alarm rate. SAM 3 with visual prompts however outperformed SAM 2 in mIoU for segmenting the sclera (likely because its segmentation masks included less of the iris). While SAM 3 furthermore appears to outperform SAM 2 in terms of miss rate, this finding has to be placed in the context of SAM 3's (much) lower ability to discriminate whether a given feature is present in an eye image, as also underscored by a very high false alarm rate for especially the iris and sclera. Given that SAM 2 outperforms SAM 3 with visual prompts while also providing a higher inference speed, we conclude that among the family of Segment Anything Models, SAM 2 remains the best choice for zero-shot eye image segmentation tasks. SAM 3's concept prompt mode does not offer a benefit over the other models for eye image segmentation.

For SAM 2, the performance reported here is in line with previous examinations of its eye image segmentation capabilities both on high-quality lab datasets \citep{niehorster2025sam2} and for the pupil in unconstrained settings \citep{maquiling2025sam2}. Regarding SAM 2 performance in in-the-wild settings, this work extends the results from \citep{maquiling2025sam2} to segmentation of the iris, sclera and entire eye opening and shows that with adequate prompting, SAM 2 can in many cases show strong performance also for these more complex eye features. 

It came as a surprise, and disappointment, to the authors that SAM 3 did not perform better than SAM 2 for eye image segmentation even when using only visual prompts. Given that SAM 3 is only 2 months old, only little work is available evaluating its performance. Nonetheless, the literature that is available indeed indicates that SAM 3 only in some cases offers superior segmentation performance in domains such as medical images \citep{chakrabarty2025comparingsam,dong2025benchmarkingsam3}, agriculture \citep{sapkota2025apples} and remote sensing \citep{li2025remotesense}. Even though SAM 3 can perform the same prompted visual segmentation task as SAM 2, the model architecture supporting this task has changed as well as its training objectives \citep[see][for a detailed discussion]{sapkota2025sam2sam3gap}. Importantly, the multimodal Perception Encoder \citep{bolya2025perceptionencoder} is used as the image encoder in SAM 3, replacing the vision-only HieraDet image encoder \citep{ryali2023hiera,bolya2023windowattention}. Furthermore, while SAM 2 was trained only to minimize geometric losses, i.e., determining where an object is, SAM 3 is trained to perform multiple tasks simultaneously, enabling it to indicate what semantic concept an image region represents \citep{sapkota2025sam2sam3gap}. While the changed architecture and training objectives extend SAM 3's capabilities, it appears that these additions do not benefit our specific visual segmentation task.

SAM 3's inferior zero-shot performance, and complete inability to provide a segmentation given the concept prompts ``iris'' and ``sclera'' may be resolved through fine tuning on a set of eye images with ground truth segmentation masks and semantic labels. Indeed, in the medical image domain, it has already been reported that major improvement in SAM 3's performance can be achieved through fine tuning \citep{jiang2026medicalsam3}. Other model adaptation strategies have also shown promising improvements in performance in medical image and camouflaged object segmentation \citep{chen2025sam3adapter}. As such, several avenues are available to potentially improve the performance of SAM 3 in eye image segmentation. We think such endeavors would be worthwhile given the promise of concept segmentation to enable eye image segmentation without any manual intervention such as the need to provide prompts for a given video. Another bottleneck in applying SAM 3 in our domain is its poor computational performance. In its current state, we found that SAM 3's inference throughput may drop below 1 fps even when using expensive workstation-class computer resources. Ongoing distillation efforts aiming to make SAM 3 more efficient \citep{zeng2025efficientsam3} may offer a solution to this performance bottleneck. Finally, SAM 3 can only be prompted using simple noun phrases. Depending on the eye images one wishes to segment, this may limit SAM 3's applicability. For instance, prompts such as ``the pupil of the left eye'' or ``the white reflection on the eye closest to the pupil'' cannot be used with vanilla SAM 3. Adaptations to SAM 3 that are able to process more complex instructions \citep{li2025sam3instructions} may offer a solution to this problem.

\subsection{Privacy and Ethics}
Like for SAM 2, the primary concern for using SAM 3 in an eye tracking context is that it is not capable of online performance. As such, eye images that are to be segmented first need to be stored after acquisition, raising privacy and data protection concerns \citep{niehorster2025sam2}. Taking a wider view, foundation models such as the SAM family partially alleviate the privacy concerns of traditional models that require large amounts of manually annotated data to be trained for a specific task. Given its strong zero-shot performance, SAM can be used for eye image segmentation without requiring training on large open datasets, while finetuning would only require a small amount of labeled data. The reduced reliance on labeled data alleviates concerns about participant privacy \citep{zhang2024medicalfoundation}.

\begin{acks}
	The authors gratefully acknowledge the Lund University Humanities Lab.
\end{acks}

\bibliographystyle{ACM-Reference-Format}
\bibliography{main}

\end{document}